\documentclass[10pt,twocolumn,letterpaper]{article}

\usepackage[pagenumbers]{cvpr} %
\usepackage{marvosym}

\usepackage{lineno} %
\usepackage{siunitx} %

\usepackage{epsfig}
\usepackage{graphicx}
\usepackage{comment}
\usepackage{amsmath,amssymb} %
\usepackage[export]{adjustbox}
\usepackage{bbm}

\usepackage[font=small,skip=5pt]{caption}
\usepackage{float}

\usepackage{multirow}
\usepackage{tabularx}
\usepackage{soul}
\usepackage{array}
\usepackage{comment}
\usepackage{outlines}

\usepackage[mathscr]{euscript}
\newcolumntype{L}[1]{>{\raggedright\let\newline\\arraybackslash\hspace{0pt}}m{#1}}
\newcolumntype{C}[1]{>{\centering\let\newline\\arraybackslash\hspace{0pt}}m{#1}}
\newcolumntype{R}[1]{>{\raggedleft\let\newline\\arraybackslash\hspace{0pt}}m{#1}}
\newcolumntype{Y}{>{\centering\arraybackslash}X}
\usepackage{booktabs}
\usepackage{enumitem}

\usepackage{amsmath}

\newcommand{\method}{CARI4D} %
\newcommand{\netName}{CoCoNet}

\newcommand{\mat}[1]{\mathbf{#1}}
\newcommand{\set}[1]{\mathcal{#1}}
\newcommand{\vect}[1]{\mathbf{#1}}

\newcommand{\pose}[0]{\boldsymbol{\theta}}

\definecolor{ForestGreen}{RGB}{14,109,14}

\def\authorspace{\quad\quad}
\newcommand{\affmark}[1]{\textsuperscript{#1}}

\definecolor{cvprblue}{rgb}{0.21,0.49,0.74}
\usepackage[pagebackref,breaklinks,colorlinks,allcolors=cvprblue]{hyperref}

\def\paperID{***}
\def\confName{CVPR}
\def\confYear{2026}

\title{
\method{}: \underline{C}ategory \underline{A}gnostic \underline{4D} \underline{R}econstruction of Human-Object \underline{I}nteraction} %

\author{Xianghui Xie\affmark{1,2,3,4,*} \authorspace Bowen Wen\affmark{1,\Letter} \authorspace Yan Chang\affmark{1} \authorspace  Hesam Rabeti\affmark{1} \authorspace Jiefeng Li\affmark{1} \\[3pt]
\authorspace Ye Yuan\affmark{1} \authorspace  Gerard Pons-Moll\affmark{2,3,4}  \authorspace Stan Birchfield\affmark{1} \\[5pt]
\small{\affmark{1}NVIDIA \quad \affmark{2}University of T\"ubingen   \quad \affmark{3}T\"ubingen AI Center \quad \affmark{4}Max Planck Institute for Informatics}\\
\small{\affmark{*}Work done during internship at NVIDIA \quad \affmark{\Letter} Corresponding author} \\
{\small\href{https://nvlabs.github.io/CARI4D/}{https://nvlabs.github.io/CARI4D/}}
}

\begin{document}
\twocolumn[{%
\renewcommand\twocolumn[1][]{#1}%
\maketitle

\begin{center}
    \centering
    \captionsetup{type=figure}
    \vspace{-5pt}
    \includegraphics[width=1.0\textwidth]{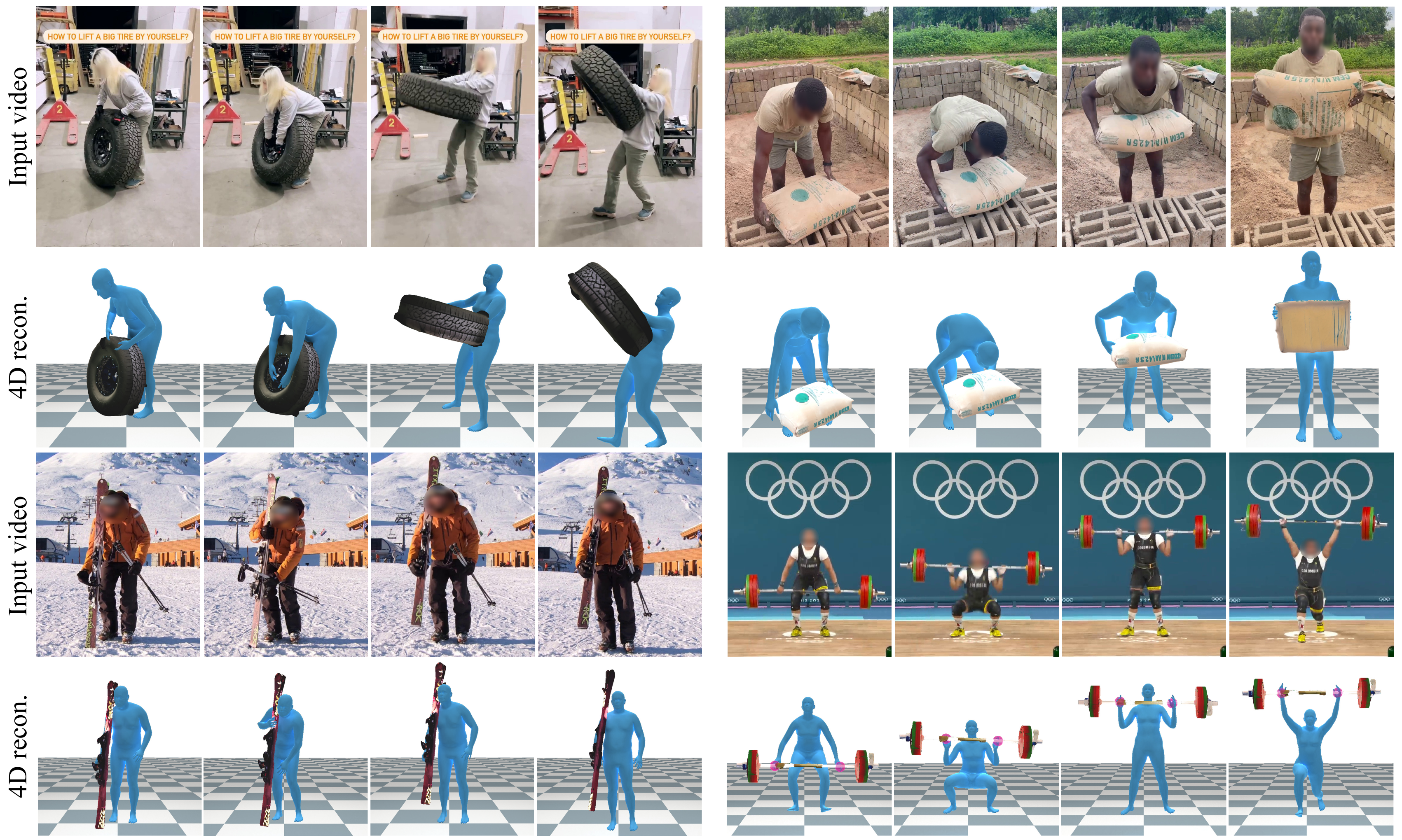}
    \vspace{-15pt}
    \captionof{figure}{\textbf{Results on in-the-wild internet videos.}
    Given a monocular RGB video, \method{} reconstructs the human and object at metric scale, and tracks the 4D human-object interaction consistently across the video. Our method is category agnostic and generalizes zero-shot.
    }
    \label{fig:teaser}
\end{center}%
}]

\begin{abstract}
Capturing human-object interaction from RGB cameras is important for human understanding, gaming, and robot learning.
Yet 4D reconstruction from a single view is hard, and prior works often assume known object template or limited categories. 
We present \method{}, the first category-agnostic method that reconstructs spatially and temporarily consistent 4D human-object interaction at metric scale from monocular RGB videos. To this end, we propose  a pose hypothesis selection algorithm that robustly integrates the individual predictions from foundation models, jointly refine them through a learned render-and-compare paradigm to ensure spatial, temporal and pixel alignment,  and finally reasoning about intricate contacts for further refinement satisfying physical constraints. 
Experiments show that our method outperforms prior art by 35\% on in-distribution dataset and 36\% on unseen dataset in terms of reconstruction error. Our model generalizes beyond the training categories and thus can be applied zero-shot to in-the-wild internet videos. Our code and pretrained models are released.

\end{abstract}
    
\section{Introduction}
\label{sec:intro}

Traditionally, capturing human-object interaction requires tedious and expensive setups, such as special studios or multi-view camera arrays~\cite{cvpr18total-capture, SIGGRAH08garment-capture, TG15free-viewpoint-video}.
While such approaches have been used in various applications (movies, gaming, robotics, augmented / virtual reality, etc.), their impact has been limited because they do not scale.

Monocular RGB videos, on the other hand, are not only abundantly available online, but new data can be captured quickly and affordably. 
As a result, approaches that can extract 4D human-object interactions directly from monocular RGB videos are highly desirable. 
Reconstruction from RGB alone, however, introduces several challenges:  (1) Both the human and object shape and pose can vary significantly between different subjects or categories; (2) The lack of depth information makes it difficult to recover scale and detailed geometry; and (3) The model needs to reason about shape, scale, pose, and dynamics while being robust to heavy occlusions.

To overcome these challenges, prior works on human-object reconstruction rely on ground truth object templates to train an instance-specific pose estimator~\cite{xie2023vistracker}, or they assume the categories are known beforehand, training models that do not generalize beyond the categories covered by the training data~\cite{xie2024InterTrack,xie2023template_free}. More recent image-based reconstruction work~\cite{cseke_tripathi_2025_pico} handles more categories by constructing databases of human-object contacts in the wild. While showing impressive results, contact retrieval is still limited to the annotated categories, and translations are inconsistent across video frames.

In this paper, we propose a method for \textbf{\underline{C}}ategory \textbf{\underline{A}}gnostic \textbf{\underline{4D}} \textbf{\underline{R}}econstruction of human-object \textbf{\underline{I}}nteraction (\textbf{\method{}}) to address these limitations. 
Our design bridges research and real-world usage by (1) reconstructing objects directly from input without pre-defined models, (2) generalizing beyond fixed categories, and (3) maintaining consistent metric reconstruction and tracking across video frames.

To solve this challenging problem, we start with existing foundation models.
Such models have shown impressive generalization on tasks like shape reconstruction~\cite{xue2024human3diffusion, hunyuan3d22025tencent, xue2024gen3diffusion, xiang2025structured3dlatents}, pose estimation~\cite{foundationposewen2024, sarandi2024nlf, genmo2025} and scene understanding~\cite{piccinelli2024unidepth, bochkovskii2025depthPro, wang2025moge}.
However, combining them is non-trivial, because their predictions lie in different coordinate spaces, they suffer from noisy input, and they do not consider the fine-grained contacts important for interaction. 
Our key idea is to carefully align the predictions from foundation models to obtain robust metric initialization, followed by an interaction-specific model that is trained to reason about contacts, which are used to further refine the interaction poses through joint optimization.

Specifically, we propose a coarse-to-fine scale estimation strategy for metric object shape reconstruction. A novel pose hypothesis selection algorithm is then used to robustly track object pose under heavy occlusion. We also adopt foundation models to obtain human pose estimations that are aligned to the same depth and scale as the object. To refine these foundation model initializations, we introduce \textbf{\netName{}}, a \textbf{\underline{C}}ategory agn\textbf{\underline{o}}stic \textbf{\underline{Co}}ntact reasoning model that renders the initial human-object estimations and compares with input observations to predict delta pose updates together with contacts. These poses and contacts are then sent into a contact-aware joint optimization framework to obtain coherent interactions. 

In summary, our main contributions are:
\begin{itemize}
    \item We present \method{}, the first category-agnostic method that reconstructs the object at metric scale and tracks the 4D human-object interaction with consistent translation and contacts, all from a single RGB video input. 
    \item We propose several novelties:  (1) A pose hypothesis selection algorithm that robustly tracks object pose under occlusions; (2) \netName{}, a category-agnostic contact reasoning network that refines human and object poses using a render and compare paradigm; and (3) a contact-aware joint optimization framework. 
    \item When trained on two real-world datasets, our method outperforms baselines by more than 36\% in terms of chamfer distance on both in-distribution and unseen datasets. Our method also generalizes to in-the-wild videos where object categories are completely unseen. 
\end{itemize}
Our code and pretrained models are publicly released.

\section{Related Works}
\noindent\textbf{Foundation models for 3D reasoning.} There has been great progress in shape reconstruction \cite{xiang2025structured3dlatents, xue2024human3diffusion, bundlesdfwen2023}, pose estimation \cite{sarandi2024nlf, foundationposewen2024}, scene understanding~\cite{wang2025vggt, dust3r_cvpr24} that reason 3D from monocular RGB cameras. High-quality object meshes with texture can be reconstructed from single image using direct 3D generation~\cite{xiang2025structured3dlatents, hunyuan3d22025tencent, zhang2024clay, TripoSR2024} or multi-view image diffusion models~\cite{xue2024gen3diffusion, xie2025MVGBench, voleti2024sv3d}. Given an object mesh,  FoundationPose~\cite{foundationposewen2024} can estimate and track the object pose consistently in an RGBD video. Similarly, recent works have enabled high-quality human avatar creation from a single image~\cite{xue2024human3diffusion, I_Ho_2024_sith}, monocular RGB videos~\cite{guo2025vid2avatarpro, reloo}, or text~\cite{xue2025infinihuman}. Large scale datasets~\cite{Black_CVPR_2023bedlam, Patel_2021_AGORA, h36m_pami} also allow training foundation models for human pose estimation~\cite{genmo2025, sarandi2024nlf, Cai_2023_SMPLerX, goel2023humans4d} that generalizes well to in-the-wild videos. For scene understanding, recent works have pushed the boundary from relative depth~\cite{wang2025moge, wang2025moge2, Ranftl2022midas, Yang2024DepthAnything} towards metric-scale depth~\cite{piccinelli2024unidepth, piccinelli2025unidepthv2, bochkovskii2025depthPro, Chen2025VideoDepthAnything} estimation and shown impressive generalization to both indoor and outdoor environments. While making significant progress, these models treat human, object, and the environment separately and do not consider their intricate interactions.

\noindent\textbf{Image-based interaction reconstruction.} Joint hand-object reconstruction has been studied for decades, with works tackling the problem with synthetic data generation~\cite{hasson19_obman}, temporal photometric consistency~\cite{hasson20_handobjectconsist}, contact potential field~\cite{yang2021cpf}, or diffusion priors~\cite{ye2023ghop, ye2023vhoi}. For full body interaction, early works~\cite{zhang2020phosa, weng2020holistic} are optimization based and follow up works propose to learn from data using distance field~\cite{xie22chore}, point cloud diffusion~\cite{xie2023template_free}, or contact transformer~\cite{nam2024contho}. Diverse interaction datasets and benchmark~\cite{xie2024rhobin} with full 3D annotation~\cite{bhatnagar22behave, huang2022intercap, zhang2023neuraldome, zhao2024imhoi, Liu_2022_CVPR_HOI4D} or contact labels~\cite{tripathi2023deco, cseke_tripathi_2025_pico, dwivedi_interactvlm_2025, zhang2024hoi, Liu_2022_CVPR, sharma_and_xie2026Hoi3DGen} are the driving force for this field. Recent efforts have pushed toward in-the-wild reconstruction for arbitrary objects~\cite{cseke_tripathi_2025_pico, dwivedi_interactvlm_2025, huo24wildhoi, wen20253dhoi} or arbitrary scenes~\cite{yalandur2025physic}. Although image-based methods demonstrate impressive generalization, they often produce temporally inconsistent results, thereby hindering downstream tasks such as robotic learning.

\noindent\textbf{Video-based interaction tracking.} Most prior works assume a known object template and focus on tracking the human and object poses. Early works track the human and object pose from multi-view RGBD~\cite{bhatnagar22behave} or RGB~\cite{sun2021HOI-FVV} cameras, and RobustFusion~\cite{robustfusion-arxiv} reduces the setup to a monocular RGBD camera. More recent work, VisTracker~\cite{xie2023vistracker}, attempts to track from monocular RGB video and handles occlusion using motion infilling. Leveraging a shape reconstruction model, InterTrack~\cite{xie2024InterTrack} can reconstruct and track the shape and pose within the trained categories. While showing strong generalization, none of these works can generalize zero-shot to unseen categories. For hand object interaction, HOLD~\cite{fan2024hold} and Diff-HOI~\cite{ye2023vhoi} are category agnostic but cannot handle large objects or heavy occlusions. Ours is the first method for full body interaction, and it generalizes zero-shot to in-the-wild videos.

\section{Method}
\begin{figure*}[t]
\centering
    \includegraphics[width=1.0\linewidth]{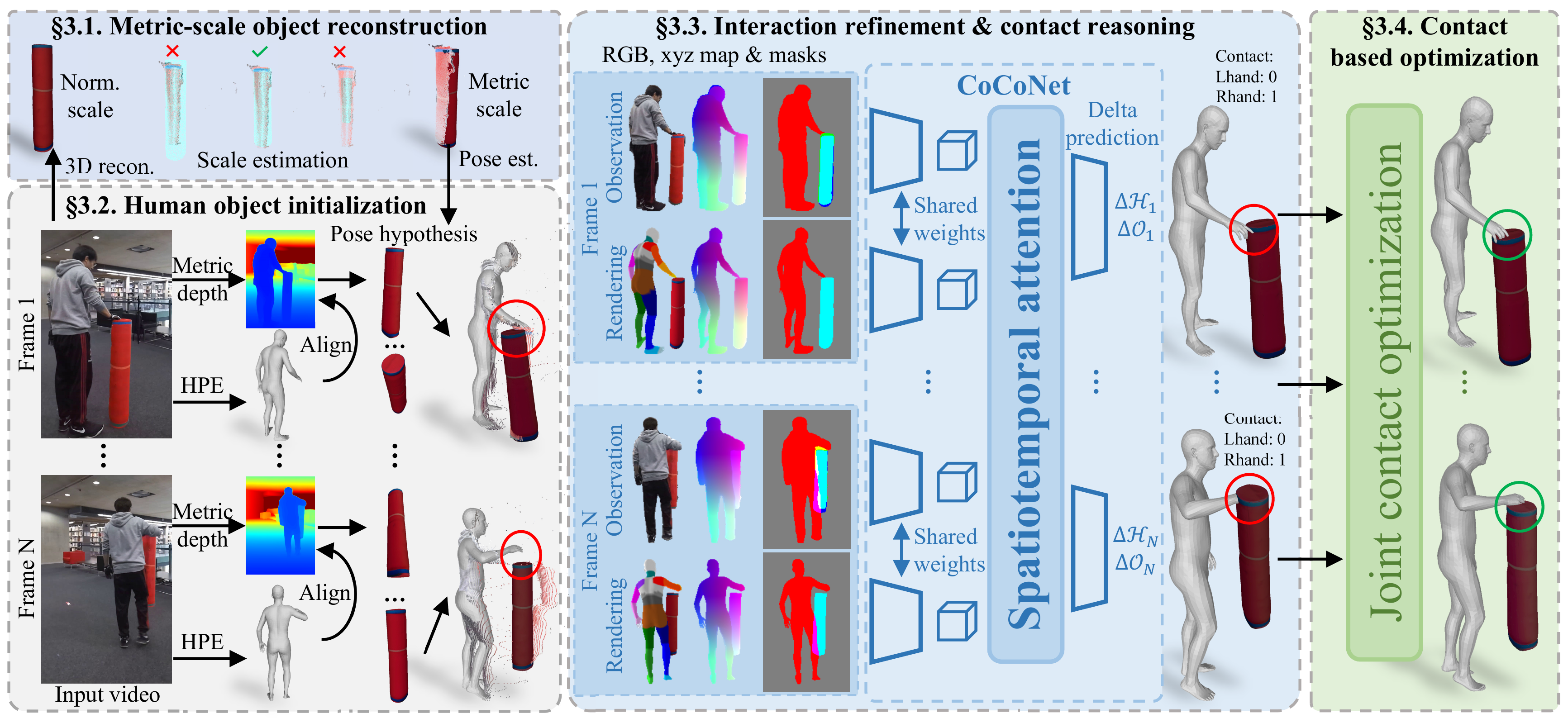}
    \vspace{-15pt}
    \caption{\textbf{\method{} method overview.} Given a monocular RGB video, we reconstruct the 4D human and object at metric scale with consistent contacts. We start by estimating the metric-scale object mesh (\cref{subsec:obj-recon}), followed by initialization of human and object poses using dynamic pose hypothesis selection (\cref{subsec:pose-init}). We then train a category agnostic contact reasoning model (\netName{}) to refine the interaction poses and estimate hand contacts (\cref{subsec:refinement}) which are used to perform contact aware joint optimization (\cref{subsec:contact-opt}).}
    \label{fig:method}
    \vspace{-10pt}
\end{figure*}

We introduce \method{}, the first category-agnostic method to reconstruct 4D human object interaction from monocular RGB video. This is a highly challenging problem due to unknown object shape, depth-scale ambiguity, dynamic motion, and severe occlusions introduced by interaction. Foundation models for shape, pose, and scene reconstruction provide strong priors yet individual predictions do not align and can suffer from noisy input. Our key idea is to design a framework that integrates the predictions to obtain a robust initialization and then design a category agnostic interaction reasoning module to improve contact coherency.

Specifically, given a sequence of $N$ RGB images $\{\mat{I}_i\}_{i=1}^N$ where a person interacts with an object, our goal is to reconstruct the object mesh $\mat{O}$ with corresponding per-frame 6DoF poses $\{\set{O}_i\}_{i=1}^N$, here $\set{O}_i=(\mat{R}^o, \vect{t}^o)$ consists of rotation $\mat{R}^o\in SO(3)$ and translation $\vect{t}^o\in\mathbb{R}^3$. For the human, we use the SMPL-H body model~\cite{smpl2015loper, MANO:SIGGRAPHASIA:2017} and estimate per-frame parameters $\{\set{H}_i\}_{i=1}^N$, where $\set{H}_i=(\pose, \beta, \vect{t}^h)$ consists of the SMPL body pose $\pose$, shape $\beta$ and global translation $\vect{t}^h$ parameters. Note that we reconstruct the human and object at \emph{metric scale} with consistent \emph{global translations} as this is important for downstream applications. 

In this section, we will first discuss how to obtain object reconstruction at metric scale (\cref{subsec:obj-recon}) and initial pose estimation (\cref{subsec:pose-init}) using foundation models. We then introduce our \netName{} 
(\cref{subsec:refinement}) to refine interaction and predict contacts, which is used to obtain more faithful interaction via joint optimization (\cref{subsec:contact-opt}). An overview of our method can be found in \Cref{fig:method}.

\subsection{Metric-scale Object Reconstruction}\label{subsec:obj-recon}
We assume that the object is mostly visible in the first frame of the video. This is common during video recordings where the human stands beside the object before actual interaction.
We hence use the first frame to obtain the object reconstruction and metric scale. Specifically, we use f-BRS~\cite{fbrs2020} to obtain object mask and run Hunyuan3D-2~\cite{hunyuan3d22025tencent} to reconstruct the object mesh. Hunyuan3D produces reasonable object mesh yet at normalized scale in its local coordinate system with unsolved pose w.r.t.\ the camera. However, we aim at obtaining metric scale 4D reconstruction. To this end, we estimate metric depth using UniDepth~\cite{piccinelli2024unidepth, piccinelli2025unidepthv2} and adopt FoundationPose~\cite{foundationposewen2024} to estimate the scale using grid search in a coarse-to-fine manner. Given a list of predefined candidate scales, we run FoundationPose to estimate the 6DoF pose for each rescaled mesh using depth and intrinsics estimated by UniDepth. We rank the candidates based on the unidirectional chamfer distance from segmented object point cloud to the mesh under the sampled scale and transformed by the estimated 6DoF pose. We then pick the top-3 scales, uniformly sample ten scales within $(0.8 s_\text{min}, 1.2 s_\text{max})$ to perform a more fine-grained scale search. 
The scale with minimum chamfer distance is used to resize the initial Hunyuan3D mesh which becomes our metric object reconstruction $\mat{O}$.

\subsection{Human and Object Pose Initialization}\label{subsec:pose-init}
Leveraging foundation models~\cite{piccinelli2025unidepthv2, sarandi2024nlf, foundationposewen2024}, we obtain robust initialization of human and object pose estimations that are aligned in metric-scale space. 

Using the metric-scale object mesh from \cref{subsec:obj-recon}, one can perform depth estimation with UniDepth~\cite{piccinelli2025unidepthv2} and solve per-frame object poses via FoundationPose~\cite{foundationposewen2024}. However, FoundationPose assumes depth from a sensor and a known textured mesh. Directly applying it with estimated depth is unreliable due to noisy depth estimation, imperfect Hunyuan3D meshes, and occlusions during interactions. To address this, we introduce a pose hypothesis selection algorithm to improve FoundationPose results in our challenging setup, producing more consistent per-frame object pose initialization for subsequent steps.

\noindent\textbf{Object pose hypothesis selection.} FoundationPose internally produces $K$ pose candidates and their quality scores per frame, selecting the pose with the highest score as the output. In our considered more challenging setup, its top-1 prediction often fails, though the correct pose typically lies within the $K$ candidates. Our key idea is to dynamically select the optimal pose per frame by leveraging visual cues and temporal consistency.
Specifically, given object pose $\hat{\set{O}}_{i-1}$ at previous frame $i-1$, our goal is to select the best pose from a list of pose hypothesis $\{\hat{\set{O}}_i^j\}_{j=1}^K$ (ranked from best to worst by FoundationPose) for current frame $i$. We consider two criteria to filter out wrong poses: a). \textbf{mask IoU} and b). \textbf{temporal smoothness.}

For \textbf{mask IoU}, we render each pose candidate $\hat{\set{O}}_i^j$ into 2D mask $\mat{M}^j_i$ and compare with object mask $\mat{M}^o_i$ from input image. We subtract the input human mask $\mat{M}_i^h$ to take occlusion into account and define the IoU as: $\text{IoU}^j_i=\frac{\sum \tilde{\mat{M}}_i^j\cap\Tilde{\mat{M}}^o_i}{\sum \Tilde{\mat{M}}_i^j\cup\Tilde{\mat{M}}^o_i}, \text{ where }\Tilde{\mat{M}}^o_i=\mat{M}^o_i\cap(\sim\mat{M}^h_i), \Tilde{\mat{M}}^j_i=\mat{M}^j_i\cap(\sim\mat{M}^h_i)$ and $\sim$ denotes mask inversion. We filter out poses whose IoU is smaller than a threshold $\delta_m$. 
For \textbf{temporal smoothness}, we compute the geodesic distance between candidate rotation $\hat{\mat{R}}^j_i$ and previous rotation $\hat{\mat{R}}_{i-1}$ and filter out poses whose distance is larger than a threshold $\delta_\text{R}$. We apply IoU filter $\delta_m$ and temporal filter $\delta_\textbf{R}$ on the $K$ pose candidates and the first one in the filtered list is the final object pose. This selection can sometimes filter out all poses due to occlusion or missing depth. In this case, we skip $S$ frames forward to a future frame where a good pose candidate can be found using criteria $\delta_m, \delta_R$ and then run pose tracking and filtering backwards. We only apply filtering if pose candidates are left during backward pose tracking and multiple forward jumps are possible in case of long term occlusion. 
To further improve the robustness, we run FoundationPose in both RGB only (setting depth image to zeros) and RGBD mode to obtain the $K$ pose candidates. 

We show in \cref{tab:ablation} that this dynamic pose selection and forward-backward pose estimation obtains significantly better result than running vanilla FoundationPose. 

\noindent\textbf{Human estimation and alignment.} For human, we run NLF~\cite{sarandi2024nlf} per-frame to obtain initial human pose estimations. NLF uses only RGB images, hence the predicted human may not align with the estimated metric depth from UniDepth~\cite{piccinelli2024unidepth} (to which object is aligned). We thus align NLF to the human depth predicted by UniDepth via optimizing the depth $z$ and a global scale of NLF prediction in an iterative closest point manner.

\subsection{Contact Reasoning and Refinement}\label{subsec:refinement}
While the pose estimations of human and object from \cref{subsec:pose-init} provides reasonable initialization, they are predicted separately and therefore do not reason about the fine-grained interactions. 
The object can be floating without contact, or penetrate with human due to noisy depth estimation or misaligned human depth, since the object and human have been treated independently in previous steps.
We hence introduce \textbf{\netName{}}, a \textbf{\underline{C}}ategory agn\textbf{\underline{o}}stic \textbf{\underline{Co}}ntact reasoning model to refine the human and object jointly while also estimating the intricate contacts. 

Specifically, given initialized human and object poses $\{\hat{\set{H}}_i, \hat{\set{O}}_i\}_{i=1}^L$ from $L$ input images (\cref{subsec:pose-init}), our \netName{} predicts the updated poses $\{\set{H}_i, \set{O}_i\}_{i=1}^L$ and binary contact labels $\{\vect{c}_i\}_{i=1}^L, \vect{c}_i\in \{0, 1\}^2$ indicating if each of the two hands is in contact with the object. Inspired by \cite{foundationposewen2024}, we design our network in a render-and-compare paradigm together with spatial temporal attention to leverage jointly the initial 3D estimation, input visual observations and temporal cues. Given our reconstructed human and object mesh, we render under their initialized poses $\hat{\set{H}}_i, \hat{\set{O}}_i$ into RGB $\hat{\mat{I}}_i$, depth $\hat{\mat{D}}_i$, and human object masks $\hat{\mat{M}}_i$. We texture the SMPL mesh with distinct vertex colors for each body part for better semantic reasoning, as shown in Fig.~\ref{fig:method} (middle). Our \netName{} $f_\phi$ then consumes a sequence of renderings $\hat{\set{I}}=\{\hat{\mat{I}}_i, \hat{\mat{D}}_i, \hat{\mat{M}}_i\}_{i=1}^L$ given the estimated pose parameters from the previous steps, and observations from the camera $\set{I}=\{\mat{I}_i, \mat{D}_i, \mat{M}_i\}_{i=1}^L$ ($\mat{D}_i$ and $\mat{M}_i$ are obtained via off-the-shelf models), and predicts delta updates $\Delta\hat{\set{H}}_i, \Delta\hat{\set{O}}_i$ for each pose parameters together with the hand contact labels $\vect{c}_i$, formally: $f_\phi: (\hat{\set{I}}(\hat{\set{H}}_i, \hat{\set{O}}_i), \set{I})\mapsto \{\Delta\hat{\set{H}}_i, \Delta\hat{\set{O}}_i, \vect{c}_i\}_{i=1}^L$. We use a frozen DINOv2~\cite{oquab2024dinov2} encoder to extract the RGB features and a trainable lightweight DINOv2 encoder to extract features from masks $\mat{M}_i$ stacked with 3D point map unprojected from the depth map $\mat{D}_i$. The encoders are shared between rendering $\hat{\set{I}}_i$ and input images $\set{I}_i$. We then apply a number of spatiotemporal attention~\cite{stabilityai2023stable} blocks on the extracted feature tokens and predict output using MLPs. 
Please refer to our Supp. for the detailed network.

\noindent\textbf{Depth alignment for \netName{} training.}
When curating the training data, our initial object poses are aligned to the predicted noisy depth~\cite{piccinelli2025unidepthv2} which deviate from the GT depth where GT poses are aligned to. 
Models trained naively 
with this mismatch tend to overfit to the error patterns of depth estimators instead of focusing on correcting the relative human-object poses. This is particularly problematic when training on multiple mixed datasets as the error patterns of metric-scale depth estimators change significantly across datasets. To this end, we propose to first align the estimated depth to GT depth, based on which, we then initialize human object poses. More specifically, we compute a scale $s$ and shift $t$ that align estimated depth $\mat{D}^\text{pr}$ to GT depth $\mat{D}^\text{gt}$: $\mat{D}^\text{align}=s\cdot\mat{D}^\text{pr} + t$. Following MiDaS~\cite{Ranftl2022midas}, we estimate $s, t$ using medians $m^\text{pr}, m^\text{gt}$ of predicted and GT depth respectively:
\begin{equation}
    \begin{aligned}
        s&=\frac{\hat{s}^\text{gt}}{\hat{s}^\text{pr}}, \quad \text{   } t=m^\text{gt} - s\cdot m^\text{pr}, \quad \text{ where} \\
        \hat{s}^\text{pr}&=\frac{\sum_i|\mat{D}^\text{pr}_i - m^\text{pr}|}{\sum_i(\mat{D}_i^\text{pr}>0)}, \quad \hat{s}^\text{gt}=\frac{\sum_i|\mat{D}^\text{gt}_i - m^\text{gt}|}{\sum_i(\mat{D}_i^\text{gt}>0)}
    \end{aligned}
\end{equation}
To improve robustness, we apply erosion and bilateral filtering to both estimated and GT depth and consider only pixels within the human and object masks to compute $s, t$. We then run FoundationPose and align the NLF prediction using the aligned depth maps. This alignment removes the absolute translation error from depth estimations and thus the network can focus on reasoning the relative poses between human and object. At test time, we do not perform any alignment to the estimated depth, since the network has learned to produce poses which are consistent with the given depth. This alignment reduces the learning efforts of our \netName{} and improves reconstruction accuracy, see \cref{tab:ablation} d and e. 

After alignment, we train our \netName{} using standard L1 loss between predicted and ground truth variables for poses $\set{H}_i, \set{O}_i$, and binary cross entropy loss for the contacts $\vect{c}_i$. 
We also annotate the symmetric transformations for the object and apply symmetry loss~\cite{wang2019normalized} for the object pose predictions. Please refer to Supp. for more details.

\subsection{Contact-based Joint Optimization}\label{subsec:contact-opt}
Our feedforward prediction from \netName{} improves the relative pose between human and object, yet it cannot guarantee the contacts are satisfied and predictions are image aligned. We further improve the contact coherency and motion smoothness via contact aware joint optimization, guided by the contact predictions from Sec.~\ref{subsec:refinement}. Specifically, we optimize the human and object pose parameters $\{\set{H}_i, \set{O}_i\}_{i=1}^N$ by minimizing the contact distances while satisfying visual constraints from input. The objective function is defined as:
\begin{equation}
    L = \lambda_c L_c + \lambda_\text{j2d} L_\text{j2d} + \lambda_m L_m + \lambda_\text{pen}L_\text{pen} + \lambda_\text{acc} L_\text{acc}
\end{equation}
where $L_c$ is the contact loss that reduces the distance between human hands to the object when contacts are predicted: $L_c = \sum_i d(\vect{J}^h_i, \mat{O}^\prime_i)\cdot \vect{c}_i$, where $d(\cdot, \cdot)$ computes the closest distance from two hand joints $\vect{J}^h_i\in \mathbb{R}^6$ to the posed object points $\mat{O}^\prime_i$. We also leverage visual cues to define 2D projection loss $L_\text{j2d}$ of SMPL body joints~\cite{xie22chore} and occlusion aware loss $L_m$ for object masks~\cite{zhang2020phosa}. We further constrain the optimization using penetration loss $L_\text{pen}$~\cite{ICCV25:VolumetricSMPL} and acceleration loss $L_\text{acc}$ that encourages the acceleration to be  zero. Exact loss definitions and weights $\lambda_*$ are discussed in Supp.

\section{Experiments}
\begin{figure*}[t]
    \centering
    \includegraphics[width=1.0\linewidth]{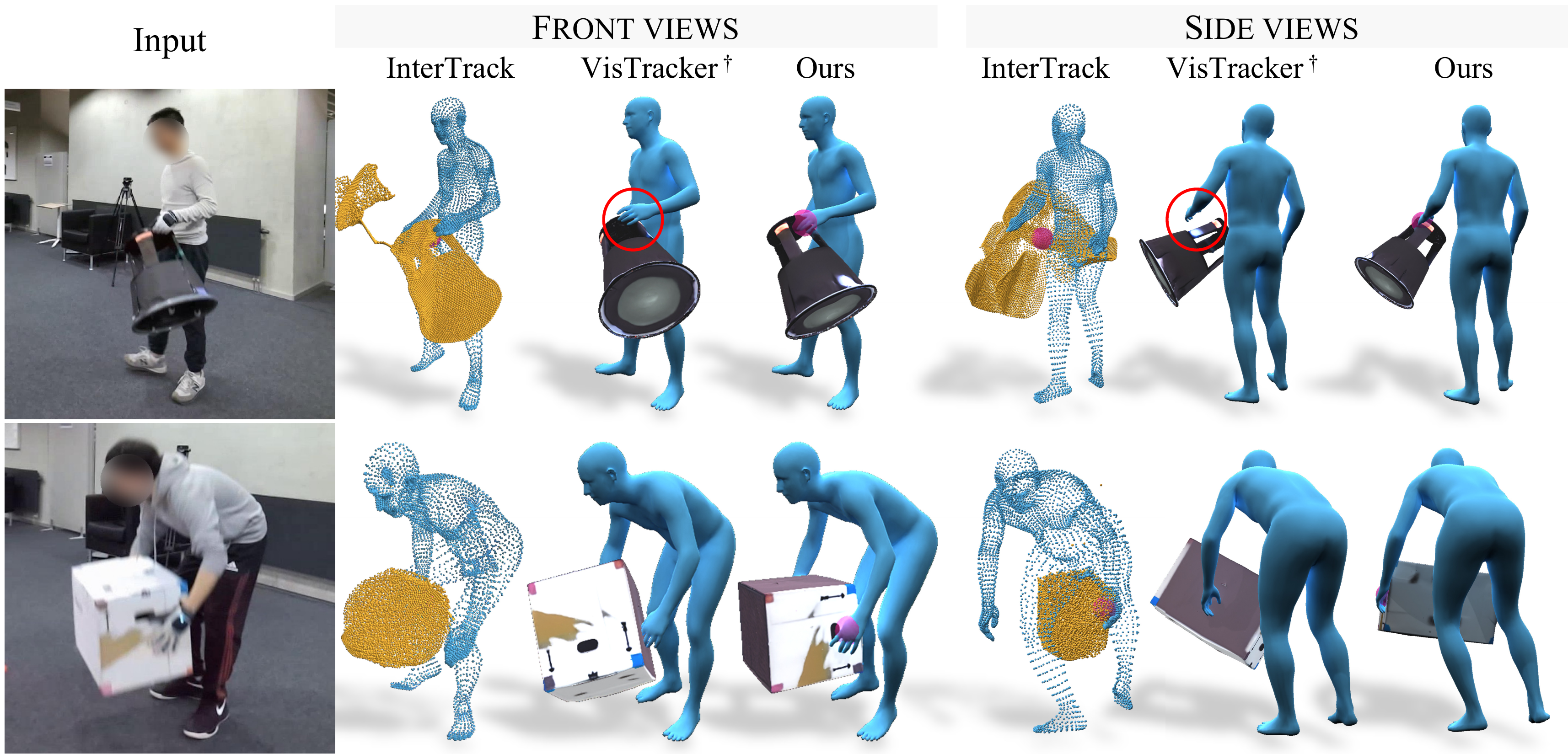}
    \vspace{-15pt}
    \caption{\textbf{Qualitative comparison on BEHAVE dataset}~\cite{bhatnagar22behave}. InterTrack~\cite{xie2024InterTrack} reconstructs human and object as point clouds only, and the shapes are noisy. VisTracker~\cite{xie2023vistracker} requires known object templates, hence we augment it with our reconstructed objects (denoted as $^\dagger$).
    Our method reconstructs the objects and tracks the poses accurately. (Purple balls indicate contact predictions.) 
    }
    \label{fig:baseline-behave}
\end{figure*}
\begin{figure*}[t]
    \centering
    \vspace{-10pt}
    \includegraphics[width=1.0\linewidth]{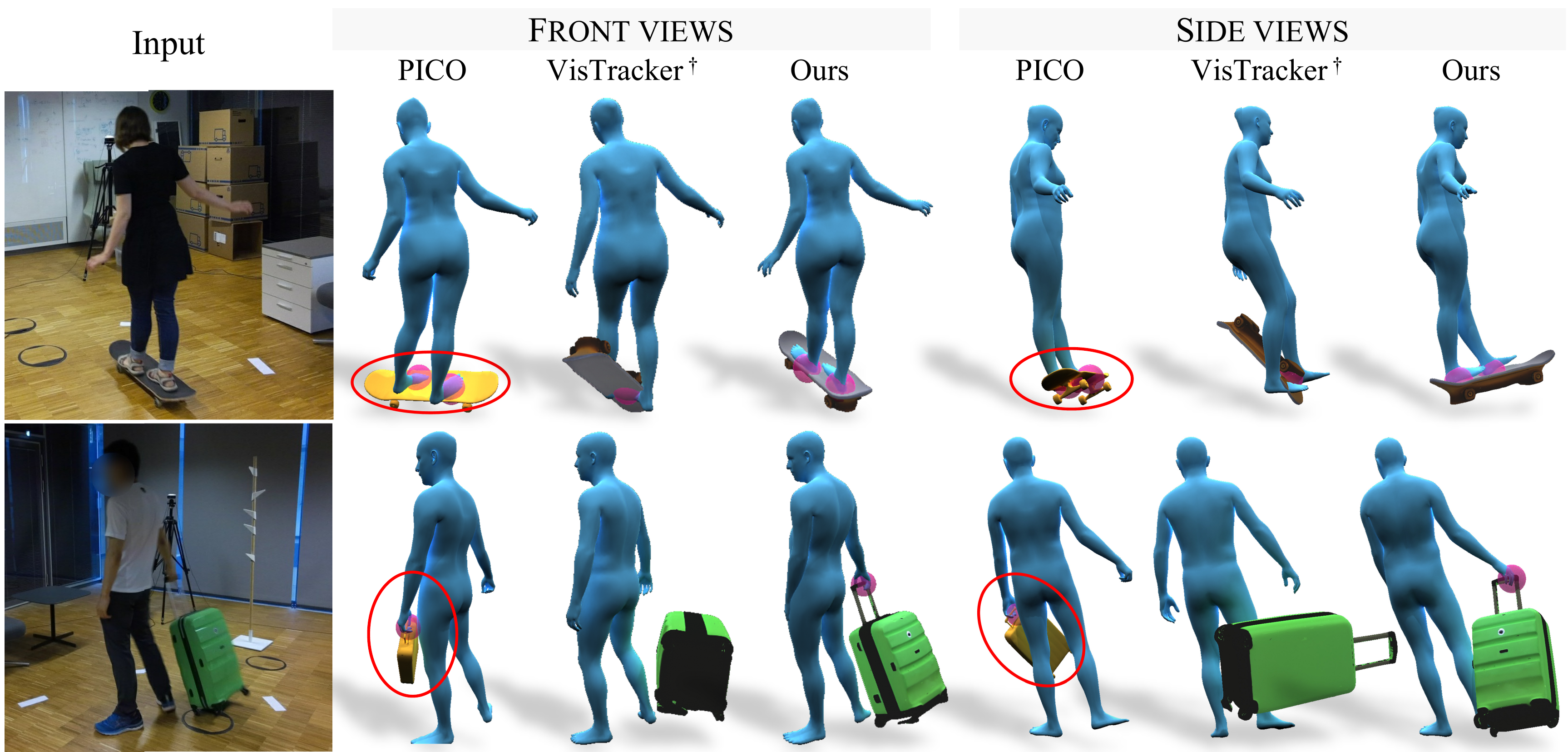}
    \vspace{-15pt}
    \caption{\textbf{Zero-shot generalization to unseen InterCap dataset}~\cite{huang2022intercap}. $^\dagger$Uses our reconstructed object meshes. Our method reconstructs the metric-scale object accurately and generalizes to unseen objects.  (Purple balls indicate contact predictions.)}
    \label{fig:baseline-intercap}
    \vspace{-10pt}
\end{figure*}
In this section, we compare our method with baselines and ablate our design choices. Results show that our method significantly outperforms all existing baselines on both in-distribution and unseen datasets. Ablation studies also validate the effectiveness of our proposed modules and our method generalizes zero-shot to in-the-wild videos. We discuss limitation and failure cases in supplementary. 

\noindent\textbf{Experiment Setup.}
We train our model on BEHAVE~\cite{bhatnagar22behave} and HODome~\cite{zhang2023neuraldome} datasets. We preprocess the training data following \cref{subsec:refinement} to align UniDepth~\cite{piccinelli2025unidepthv2}, NLF~\cite{sarandi2024nlf}, and FoundationPose~\cite{foundationposewen2024} predictions to ground truth depth. We test the models on the BEHAVE test set and unseen InterCap~\cite{huang2022intercap} test set, following the same split as prior works~\cite{xie2023vistracker, cseke_tripathi_2025_pico}. Both BEHAVE and InterCap were captured in multi-view camera setup, and previous methods simply pick a fixed camera view for test regardless of the occlusion and view angles. Although this is the most general setting, it is far from a practical video recording setup. We hence select test views where the object is mostly visible in the first few frames, %
making the evaluation more suitable to downstream applications. In total 62 videos from BEHAVE and 22 videos from InterCap are selected for evaluation. 

\noindent\textbf{Evaluation Metrics.}
We report the chamfer distance~(cm) between reconstructed and GT meshes of SMPL (CD-h), object (CD-o) and combined mesh (CD-c). Prior works perform frame based~\cite{xie22chore, cseke_tripathi_2025_pico, xie2023template_free, nam2024contho} with GT to remove the depth-scale ambiguity. This ignores the negative impact of temporally inconsistent scale or translation drift that frequently occurs in 4D reconstruction. Instead, to consider the world space translation, we align the first frame reconstruction to GT and apply the same transformation to the full video. For the first frame alignment, We utilize SMPL mesh only, instead of the combined human object mesh (\cite{xie2023vistracker}), as the reconstructed objects do not have the same topology as the GT mesh. We also report the acceleration error to measure the smoothness of the reconstructed motion. We define human acceleration error (Acc-h) as the mean per joint acceleration error against ground truth~\cite{zeng2022smoothnet, genmo2025} and object acceleration error (Acc-o) as the translation acceleration against GT.

\subsection{Baseline Comparison}
We compare against InterTrack~\cite{xie2024InterTrack} and VisTracker~\cite{xie2023vistracker} on the BEHAVE test set in \cref{tab:comp-behave}. All models have been trained with BEHAVE training set. Template-free approach, InterTrack, reconstructs human and object as 4D point cloud sequences normalized at each frame. Its lack of surface reconstruction and varying scale across frames makes it difficult for downstream applications. Furthermore, the translation from InterTrack is estimated using heuristics and 2D bounding boxes, leading to inconsistent global translations and thereby much higher errors than other methods under the 4D globally aligned setup
(\cref{tab:comp-behave} row 1). VisTracker requires known object templates as input. 
In order to adapt to the considered setup where object information is unknown, we align our reconstructed object meshes with GT and augment VisTracker with the aligned object reconstructions. While producing consistent translation and smooth motion, VisTracker predicts less accurate results due to noisy pose and contact predictions. 
Our method, on the other hand, not only reconstructs the object meshes more accurately with metric scale, but also recovers the human and object poses coherently across the full video, outperforming prior methods by over 35\% in terms of chamfer distance (CD-c). We also show some qualitative comparison in \cref{fig:baseline-behave}.

\begin{table}[ht]
    \centering
    \footnotesize
    \begin{tabular}{l|>{\centering\arraybackslash}p{0.8cm} >{\centering\arraybackslash}p{0.8cm} >{\centering\arraybackslash}p{0.8cm} >{\centering\arraybackslash}p{0.9cm} >{\centering\arraybackslash}p{0.9cm}}
    \toprule[1.5pt]
        Method & CD-h$\downarrow$ & CD-o$\downarrow$ & CD-c$\downarrow$ & Acc-h$\downarrow$ & Acc-o$\downarrow$ \\
        \midrule
        InterTrack & 25.71 & 47.66 & 30.20 & 5.23 & 5.64\\ 
        VisTracker & 13.52 & 18.29 & 14.22 & {\bf 0.54} & 0.77 \\
        Ours & {\bf 7.74} & {\bf 12.05} & {\bf 9.23} & 1.14 & {\bf 0.75}\\ 
    \bottomrule[1.5pt]
    \end{tabular}
    \vspace{-5pt}
    \caption{\textbf{Evaluation results on BEHAVE}~\cite{bhatnagar22behave} dataset (unit: cm). Our method significantly outperforms previous instance-specific VisTracker~\cite{xie2023vistracker} and category-specific InterTrack~\cite{dwivedi_interactvlm_2025}. }
    \label{tab:comp-behave}
    \vspace{-10pt}
\end{table}

\subsection{Zero-shot Generalization}
\begin{figure}
    \centering
    \includegraphics[width=1.0\linewidth]{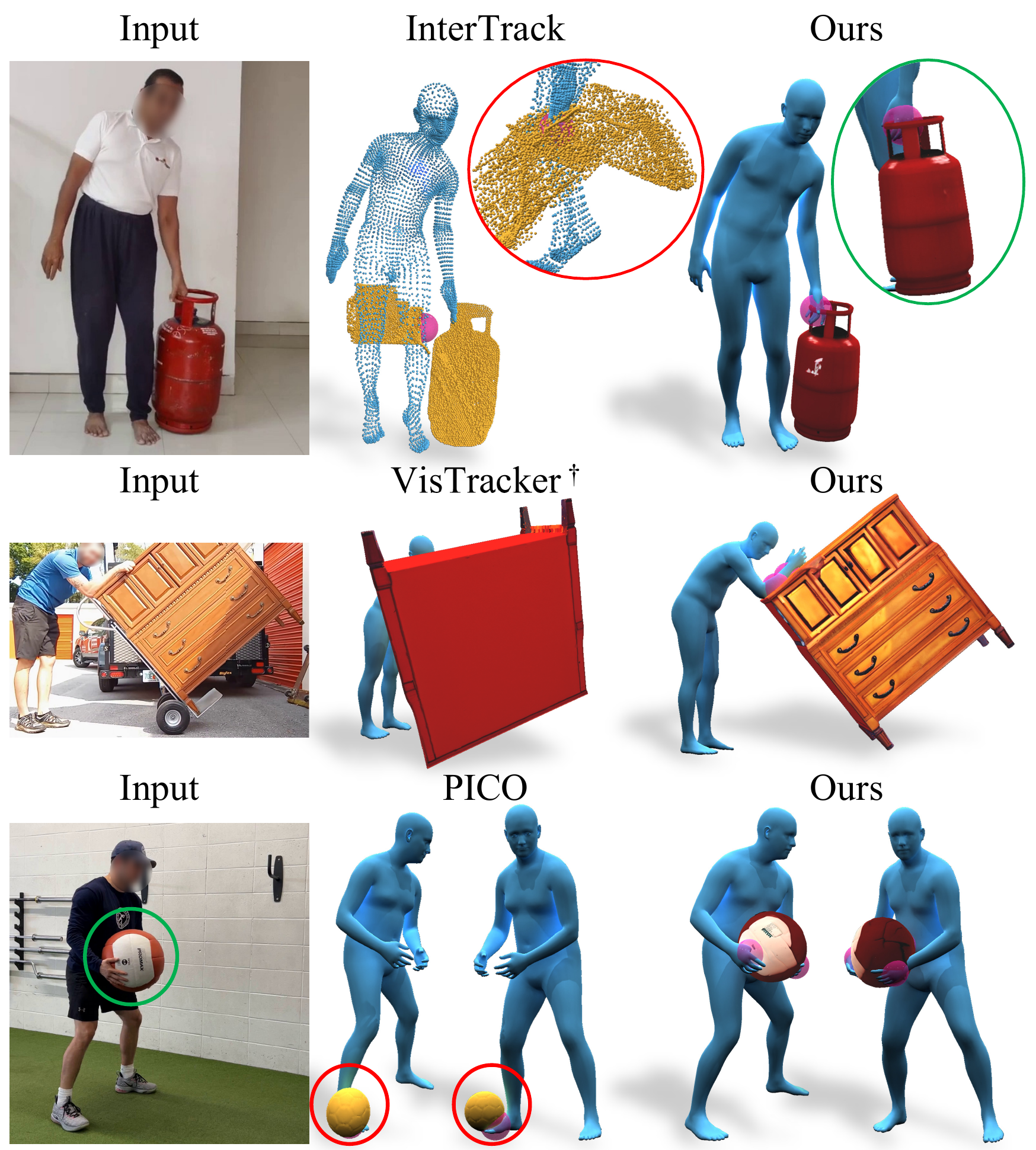}
    \vspace{-15pt}
    \caption{\textbf{Generalization to in-the-wild videos}. Prior methods predict noisy shape (InterTrack~\cite{xie2024InterTrack}), flipped object pose (VisTracker~\cite{xie2023vistracker}, $^\dagger$ with our object reconstruction) or wrong contacts and object position (PICO~\cite{cseke_tripathi_2025_pico}). Our method generalizes better overall. 
    (Purple balls indicate contact predictions.)
    }
    \label{fig:wild-compare}
    \vspace{-10pt}
\end{figure}

We evaluate the zero-shot generalization ability of our method on the InterCap~\cite{huang2022intercap} test set and compare with image based method PICO~\cite{cseke_tripathi_2025_pico} and video based tracking methods InterTrack and VisTracker~\cite{xie2024InterTrack, xie2023vistracker}. All methods are not trained on InterCap. 
PICO is image based optimization method which is extremely computationally expensive (5min/image) to run on full videos hence we compare with it on key frames (3fps) only. PICO relies on predicted contacts on human body~\cite{tripathi2023deco} to retrieve an object mesh and one-to-one contact correspondence between human and object vertices. Due to noisy human contact predictions from ~\cite{tripathi2023deco}, its retrieval may result in wrong correspondence in the meshes such as swapped left and right (\cref{fig:baseline-intercap} row 2) or wrong contact location (\cref{fig:wild-compare} row 3), leading to large errors in both object pose and shape. 

InterTrack and VisTracker are category or instance specific tracking methods, hence struggle to generalize to unseen instances or categories in InterCap. In contrast, our method is category agnostic and generalizes to novel instances and categories, see also the qualitative comparison in \cref{fig:baseline-intercap}. We also test our method on in-the-wild internet videos, and results are shown in \cref{fig:teaser} and \cref{fig:wild-compare}. Our method generalizes well to diverse object shapes, categories and interaction types. We show more results in Supp. video. 

\begin{table}[ht]
    \centering
    \footnotesize
    \begin{tabular}{l>{\centering\arraybackslash}p{0.8cm} >{\centering\arraybackslash}p{0.8cm} >{\centering\arraybackslash}p{0.8cm} >{\centering\arraybackslash}p{0.9cm} >{\centering\arraybackslash}p{0.9cm}}
    \toprule[1.5pt]
        Method & CD-h$\downarrow$ & CD-o$\downarrow$ & CD-c$\downarrow$ & Acc-h$\downarrow$ & Acc-o$\downarrow$ \\
        \midrule
        PICO$^*$& 5.15  & 27.63  &  87.73 & - & - \\ 
        \textbf{Ours$^*$} & {\bf 3.21  } & {\bf 9.04 }& {\bf 5.90  } & - & - \\ 
        \hline  %
        InterTrack & 24.83 & 35.88 & 27.69 & 4.73 & 4.79 \\ 
        VisTracker & 16.12 & 27.41 & 20.17 & {\bf 0.98} & 1.25\\
        \textbf{Ours} & {\bf 11.06} & {\bf 15.69} & {\bf 12.88} & 1.25 & {\bf 0.82} \\ 
    \bottomrule[1.5pt]
    \end{tabular}
    \caption{\textbf{Zero-shot generalization to unseen InterCap}~\cite{huang2022intercap} dataset (unit: cm). $^*$Denotes key-frames only, where acceleration metrics do not apply. Our method outperforms both image based method PICO~\cite{cseke_tripathi_2025_pico} and video based tracking methods~\cite{xie2024InterTrack, xie2023vistracker}.}
    \label{tab:comp-intercap}
\end{table}

\subsection{Ablation Studies}
We ablate the design choices of our method using a subset (24 videos, cover all objects) of the full BEHAVE~\cite{bhatnagar22behave} test set. We report chamfer and smoothness errors in \cref{tab:ablation}. 

\noindent\textbf{Human and object initialization.} We propose a dynamic pose hypothesis selection algorithm to obtain more robust object pose tracking using FoundationPose (FP~\cite{foundationposewen2024}). We also align NLF~\cite{sarandi2024nlf} human predictions to UniDepth~\cite{piccinelli2025unidepthv2} depth estimation such that human and object are in the same metric-scale space. In \cref{tab:ablation}, we compare our proposed initialization (row c) against the raw NLF predictions (w/o alignment) combined with FP in tracking (row a) or per-frame estimation (row b) mode. Our initialization is significantly better than raw NLF and FP results.

\sisetup{
    mode = text,                 %
  detect-weight=true,
  detect-inline-weight=math,
  table-number-alignment=right
}

\begin{table}[ht]
  \centering
  \scriptsize
  \begin{tabular}{
    l|
    S[table-format=2.2,table-column-width=0.5cm]
    S[table-format=4.2,table-column-width=0.6cm]
    S[table-format=3.2,table-column-width=0.6cm]
    S[table-format=1.2,table-column-width=0.5cm]
    S[table-format=1.2,table-column-width=0.5cm]
  }
    \textbf{Method} &
    {\tiny CD-h$\downarrow$} &
    {\tiny CD-o$\downarrow$} &
    {\tiny CD-c$\downarrow$} &
    {\tiny Acc-h$\downarrow$} &
    {\tiny Acc-o$\downarrow$} \\
    \hline
    a. Raw NLF + FP tracking       & 11.53 & 1565.42 & 405.13 & 3.06 & 4.34 \\
    b. Raw NLF + FP pose est.      & 11.53 & 40.54   & 13.02  & 3.06 & 9.26 \\
    c. Our init. NLF + FP          & 7.81  & 16.85   & 10.79  & 2.21 & 8.36 \\
    d. Our init. + ref. w/o align  & 8.01  & 13.55   & 9.95   & 2.08 & 3.74 \\
    e. Our init. + ref. w/ align   & \bfseries 7.01 & 11.59 & \bfseries 8.62 & 1.75 & 3.78 \\
    f. Our full model              & 8.41 & \bfseries 11.57 & 9.35 & \bfseries 1.06 & \bfseries 0.38 \\
  \end{tabular}
  \caption{\textbf{Ablation studies.} Our proposed initialization (c) is better than running vanilla (a, b) NLF~\cite{sarandi2024nlf} and FoundationPose (FP~\cite{foundationposewen2024}). Our contact reasoning model trained with the proposed alignment (e) further improves the accuracy. Joint optimization (f) improves smoothness and contact consistency (see \cref{fig:ablation}).}
  \label{tab:ablation}
\end{table}

\noindent\textbf{Contact reasoning model \netName{} and training.} We introduce \netName{} to further refine our initialized human and object poses and reason about the contacts. In \cref{tab:ablation} we show that our \netName{} (row e) improves our initial human and object reconstructions (row c). 
To ensure consistency between the ground-truth poses from the datasets and the initialized metric-scaled human, object, and monocular depth, we propose to align them for training. To ablate this training strategy, we then train a model without our alignment (row d) which yields even worse  results than the initialization on human, since the model is tasked for an ill-posed problem: predicting the correct pose updates for the misaligned human pose and monocular depth.
This highlights the importance of our \netName{} and training strategy.

\begin{figure}[t]
    \centering
    \includegraphics[width=1.0\linewidth]{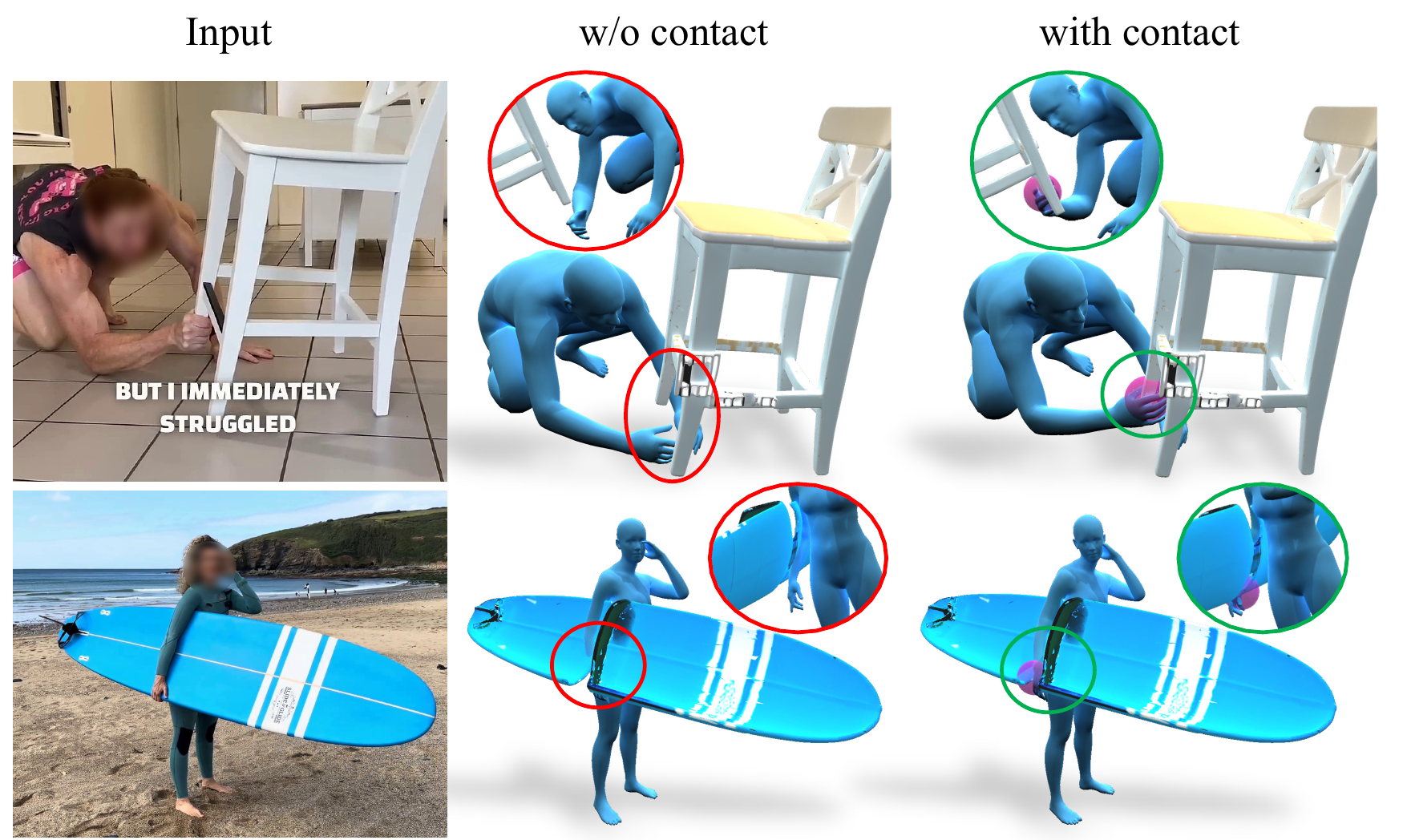}
    \vspace{-15pt}
    \caption{\textbf{Importance of contacts}. Without our contact-aware optimization, the model does not properly handle the fine-grained hand-object interaction, leading to floating object or penetration errors. (Purple balls indicate contact predictions.) %
    }
    \label{fig:ablation}
\end{figure}

\noindent\textbf{Contact based joint optimization.} Our \netName{} predicts contacts which is then used to further optimize the output poses from our \netName{} to satisfy physical constraints. Compared to the initial prediction from \netName{} (\cref{tab:ablation} e), our joint optimization improves the motion smoothness and coherency of the contacts. We show two examples in \cref{fig:ablation}. Without optimization, the model is not aware of the intricate contact points on the object, leading to artifacts such as penetration or floating objects. Our optimization refines the object and hand poses, leading to smoother and more coherent reconstruction (\cref{tab:ablation} between f and e). 

\noindent\textbf{Oracle study.}
The primary focus of this paper is to develop a general method for RGB based 4D reconstruction without any object shape or depth information as inputs. We test the performance of our method when their privileged information is given in \cref{tab:gt-input}. It can be seen that ground truth object shape or depth improves slightly the object reconstruction and combining both achives the most boost. Our base model designed for RGB input is close to the upper bound with oracle GT information.

\begin{table}[ht]
    \centering
    \footnotesize
    \begin{tabular}{l|>{\centering\arraybackslash}p{0.7cm} >{\raggedleft\arraybackslash}p{0.7cm} >{\centering\arraybackslash}p{0.7cm} >{\centering\arraybackslash}p{0.75cm} >{\centering\arraybackslash}p{0.75cm}}
        Method & {\scriptsize CD-h$\downarrow$} & {\scriptsize CD-o$\downarrow$} & {\scriptsize CD-c$\downarrow$} & {\scriptsize Acc-h$\downarrow$} & {\scriptsize Acc-o$\downarrow$} \\
        \hline
        
        Ours & 8.41    & 11.57    & 9.35    & 1.06    & 0.38  \\ 
        +GT mesh & 8.05    & 9.26    & 8.18    & 1.03    & 0.32 \\
        +GT dmap & 7.50    & 11.25    & 8.78    & 0.94    & 0.35  \\ 
        +GT mesh\&dmap & 7.23    & 7.78    & 7.20    & 0.92    & 0.27  \\ 
    \end{tabular}
    \vspace{-5pt}
    \caption{\textbf{Oracle study}. Trained on estimated depth, our model also allows input with ground truth depth or object mesh and achieves slightly better results.}
    \label{tab:gt-input}
    \vspace{-10pt}
\end{table}

\section{Conclusion}
We present \method{}, a category agnostic method to reconstruct 4D human object interaction from monocular RGB video. Our key idea is to design a framework that aligns the predictions from foundation models to obtain a robust initialization followed by learned contact reasoning and interaction refinement. We first perform coarse to fine scale estimation to recover metric-scale for the object and then introduce pose hypothesis selection for robust object pose tracking. We then train our \netName{} that predicts hand contacts and refines the initial interaction poses which is then used to jointly optimize the poses together with contacts. We evaluate \method{} on BEHAVE and unseen InterCap datasets, achieving over 35\% improvement in Chamfer distance for both in-distribution and zero-shot settings. Ablations confirm the effectiveness of our pose selection, \netName{}, and other design choices. \method{} also generalizes well to in-the-wild videos with diverse objects and interactions. Code and pretrained models are released.

{
    \small
    \bibliographystyle{ieeenat_fullname}
    \bibliography{main}
}

\clearpage
\setcounter{page}{1}
\maketitlesupplementary

In this supplementary, we discuss in more details about our \netName{} architecture, joint optimization and runtime performance. We then discuss limitations and failure cases. Please refer to our supplementary video for 4D reconstruction results and baseline comparisons. 

\section{Implementation Details}\label{sec:implementation}
We detail our implementations in this section. Note that our code and pretrained models will be fully released with detailed documentation to enable reproduction of our results. 
\subsection{\netName{} Details}\label{subsec:coconet}
\begin{figure*}[t]
    \centering
    \includegraphics[width=1.0\linewidth]{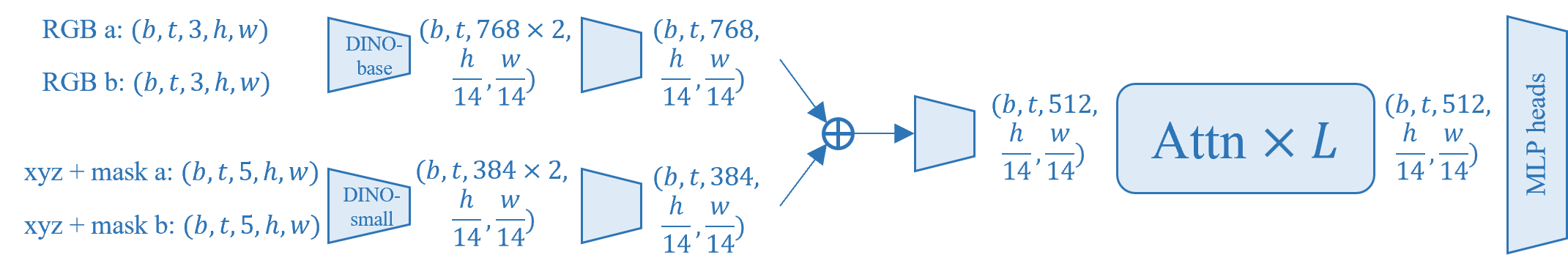}
    \caption{\textbf{\netName{} architecture}. Here $b, t, h, w$ denote batch size, temporal window size, image height and width respectively. We follow a render-and-compare paradigm, hence RGB $a$ and RGB $b$ denote the image from input observation and rendering respectively, same for xyz map and mask (human and object stacked together). }
    \label{fig:coconet-detail}
\end{figure*}

\paragraph{Network architecture.} We plot the architecture diagram of our \netName{} in \cref{fig:coconet-detail}. We adopt DINOv2~\cite{oquab2024dinov2} as our image encoder and keep the base DINO for RGB image frozen while the small DINO model for xyz map and mask is fine tuned. We replace the first convolution layer in DINOv2 small to consume five channels (xyz + human object masks) instead of three RGB channels and finetune it end to end with other network layers. The attention block is spatiotemporal attention module similar to SV3D~\cite{voleti2024sv3d}. Specifically, given feature grid of shape $(b, t, d, h, w)$, we first reshape and perform spatial attention on feature $(bt, hw, d)$ to obtain $\mat{F}_s$, followed by a temporal attention on feature $(bhw, t, d)$ to obtain $\mat{F}_t$. We then blend spatial feature $\mat{F}_s$ and temporal feature $\mat{F}_t$ with a learnable factor $\alpha\in [0, 1]: \mat{F} = (1-\alpha)\mat{F}_s + \alpha \mat{F}_t$. We use two spatiotemporal attention blocks in our network and temporal window size is $t=96$. Following the spatiotemporal attention are five MLP heads that regress delta updates of object rotation $\Delta \mat{R}\in \mathbb{R}^6$, object translation $\Delta \mat{t}^o\in \mathbb{R}^3$, SMPL pose $\Delta \pose \in \mathbb{R}^{144}$, SMPL shape $\Delta \beta \in \mathbb{R}^{10}$, and SMPL translation $\Delta \vect{t}^h \in \mathbb{R}^3$. 

\paragraph{Training details.} We train our model on BEHAVE~\cite{bhatnagar22behave} and HODome~\cite{zhang2023neuraldome} dataset following the data preprocessing discussed in \cref{subsec:refinement}. We use AdamW optimizer with a learning rate of 1e-4 and an effective batch size of 16 for 38k steps. The training process takes about 38 hours on 8$\times$A100$@$80GB GPUs. 
\subsection{Joint Optimization}\label{subsec:opt-details}
We discuss in more details the objective function for our contact aware joint optimization (\cref{subsec:contact-opt}). Given a sequence of refined human-object poses $\{\set{H}_i, \set{O}_i\}_{i=1}^N$ and binary hand contact labels $\{\vect{c}_i\}_{i=1}^N$ predicted by our \netName{}, we aim at improving its contact realism and avoiding penetration via optimizing the pose parameters $\{\set{H}_i, \set{O}_i\}_{i=1}^N$. 
The overall objective function is defined as:
\begin{equation}
    L = \lambda_c L_c + \lambda_\text{j2d} L_\text{j2d} + \lambda_m L_m + \lambda_\text{pen}L_\text{pen} + \lambda_\text{acc} L_\text{acc}
    \label{eq:supp-opt-loss}
\end{equation}
where $L_c, L_\text{j2d}, L_m, L_\text{pen}, L_\text{acc}$ are the contact loss, 2D body joint reprojection loss, object mask loss, penetration loss and acceleration loss respectively. 

The contact loss $L_c$ penalizes large distances between hand and object when there is contact: $L_c = \sum_i d(\vect{J}^h_i, \mat{O}^\prime_i)\cdot \vect{c}_i$, where $d(\cdot, \cdot)$ computes the closest distance from two hand joints $\vect{J}^h_i\in \mathbb{R}^6$ to the posed object points $\mat{O}^\prime_i$. 

The 2D joint projection loss $L_\text{j2d}$ minimizes the distance between projected 2D joints $\pi(\mat{J}(\mat{H}_i))$ and detected 2D joints $\hat{\mat{J}}_i$ from input image using openpose~\cite{openpose}: $L_\text{j2d}=\sum_i||\hat{\mat{J}}_i-\pi(\mat{J}(\mat{H}_i))||_2^2$, where $\mat{J}(\cdot)$ regresses the 3D body joints from SMPL mesh $\mat{H}_i$ and $\pi(\cdot)$ projects the 3D joints to 2D image. 

The 2D object mask loss $L_m$ is occlusion aware and defined as: $L_m=\sum (M - I\circ S)^2$, where $M$ is the input object mask, $I$ is a non-occlusion indicator (1 if pixel belongs only to this object, 0 else) and $S$ is the rendered object silhouette. The non-occlusion indicator is derived from human and object masks. The human and object masks may have overlap regions due to imperfect segmentation. To address this, pixels that belong only to object mask are assigned value 1 for the non-occlusion indicator.
This avoids computing loss on regions where the object is occluded, see illustration in PHOSA~\cite{zhang2020phosa}. 

We define the penetration loss using Volumetric SMPL~\cite{ICCV25:VolumetricSMPL} which learns a function $\Phi_\text{SMPL} $ that predicts a signed distance given a query point $q\in \mathbb{R}^3$, formally: $\Phi_\text{SMPL}: \mathbb{R}^3\mapsto \mathbb{R}$. The penetration loss is hence defined as: $L_\text{pen}=\sum_i\sum_{q\in\mat{O}_i^\prime}\text{ReLU}(-\Phi_\text{SMPL}(q))^2$. In practice we sample 6000 points on the object mesh surface as query points to compute the penetration loss.  

The acceleration loss $L_\text{acc}$ avoids large jitters by pushing the acceleration to be zero. In general: $L_\text{acc}=||\vect{x}_{i}-2\vect{x}_{i-1}+\vect{x}_{i-2}||_2^2$, here $\vect{x}$ includes 3D human body joints $\mat{J}(\mat{H})$ and object poses. 

The loss weights are $\lambda_c=200,\lambda_\text{j2d}=0.03, \lambda_m=0.002, \lambda_\text{pen}=2.0$ and we use $\lambda_\text{acc}=600$ for human body joints, $\lambda_\text{acc}=1000$ for object poses. We use Adam optimizer with linear learning rate decay (start with 1e-3 learning rate) to optimize the parameters for 3000 steps. For efficiency we add penetration loss only in the last 1200 steps. 

\subsection{Runtime Performance}
We report the average runtime of different methods to finish processing a video sequence of 300 frames using one A100$@$80GB GPU in \cref{tab:runtime}. Image based approach PICO~\cite{cseke_tripathi_2025_pico} requires the longest time as it optimizes one image each time and each optimization takes almost five minutes to finish. Video based methods InterTrack~\cite{xie2024InterTrack} and VisTracker~\cite{xie2023vistracker} are faster as they leverage temporal cues and process multiple frames in parallel. However, they are still lower than our method due to complex multi-stage optimization (VisTracker~\cite{xie2023vistracker}) or diffusion sampling process (InterTrack~\cite{xie2024InterTrack}). Our method achieves the fastest runtime while still producing the most accurate results. 

\begin{table}[h]
    \centering
    \small
    \begin{tabular}{c|c c c c}
         Method & PICO & InterTrack & VisTracker & Ours \\
         \hline
         Runtime (min) $\downarrow$& 1560 &   198 & 118 & {\bf 45}\\
    \end{tabular}
    \caption{\textbf{Average runtime (minutes) }to process one video of 300 frames. Our method is much faster than baselines while being more accurate.}
    \label{tab:runtime}
\end{table}

\section{Further Analysis}
We provide additional analysis of our method under different conditions in this section. 

\noindent\textbf{Initialization and error propagation of foundation models}. Our method relies mainly on FoundationPose(FP) and UniDepth predictions. 
UniDepth already provides accurate priors (9.20 cm avg. error on BEHAVE).
We mitigate foundation model errors via our pose hypothesis selection and CoCoNet refinement. 
We now further validate robustness by injecting noise into FoundationPose and UniDepth predictions
and run full CARI4D pipeline on BEHAVE (same test set as Tab 3). Plots below show the final chamfer distance (cm) of combined human-object mesh vs.\ injected noise magnitude, together with the error of VisTracker. Results confirm CARI4D remains stable and superior to VisTracker even under significant initialization noise, being insensitive to foundation model errors.
\begin{figure}[h]
    \centering
    \vspace{-8pt}
    \includegraphics[width=1.0\linewidth]{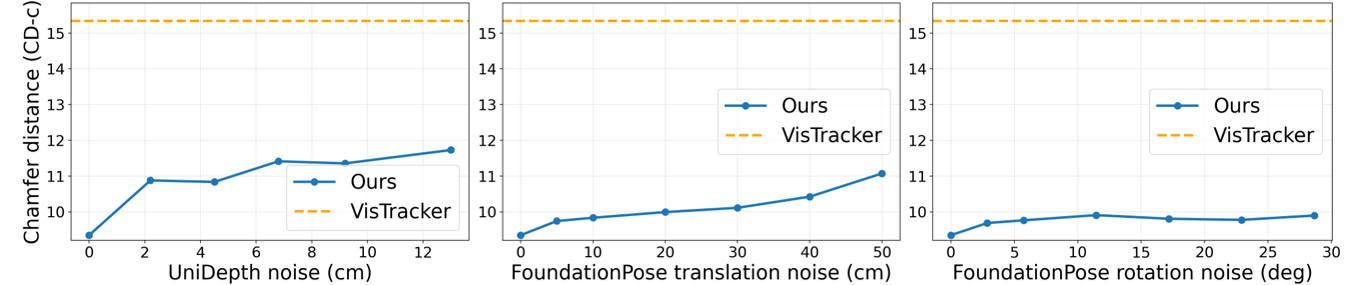}
    \caption{Sensitivity analysis of UniDepth and FoundationPose errors on the final performance.}
    \label{fig:sensitivity}
\end{figure}

\noindent\textbf{Reliability and impact of object occlusion.} We assume 1st frame is mostly visible, which reduces the hallucination for object mesh reconstruction. We plot the final object error versus the object visibility in the reconstruction frame in the figure below. There is no clear correlation between the visibility for reconstruction and final error because the performance also depends on the occlusion in other frames and difficulty of the motion. As long as the frame used to reconstruct is reasonably visible ($>86\%$ in our case), it does not significantly affect the final performance. We also provide the object mask for mesh reconstruction and pose tracking to identify target object. 
\begin{figure}[h]
    \centering
    \includegraphics[width=0.9\linewidth]{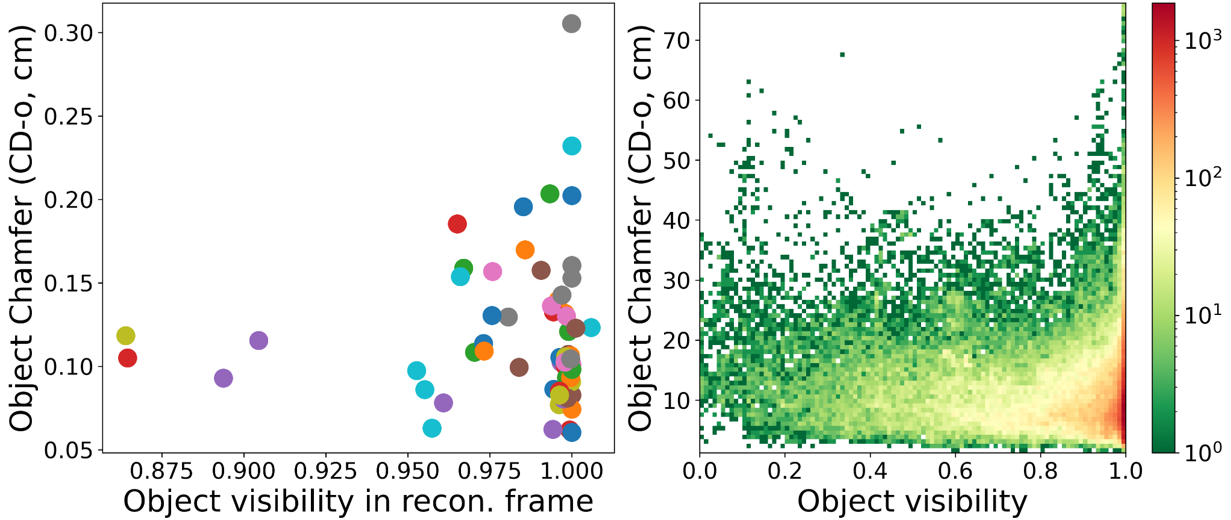}
    \caption{Error per seq. (left) and visibility distribution (right).}
    \label{fig:error-vs-occlusion}
    \vspace{-6pt}
\end{figure}

\begin{figure}[h]
    \centering
    \includegraphics[width=1.0\linewidth]{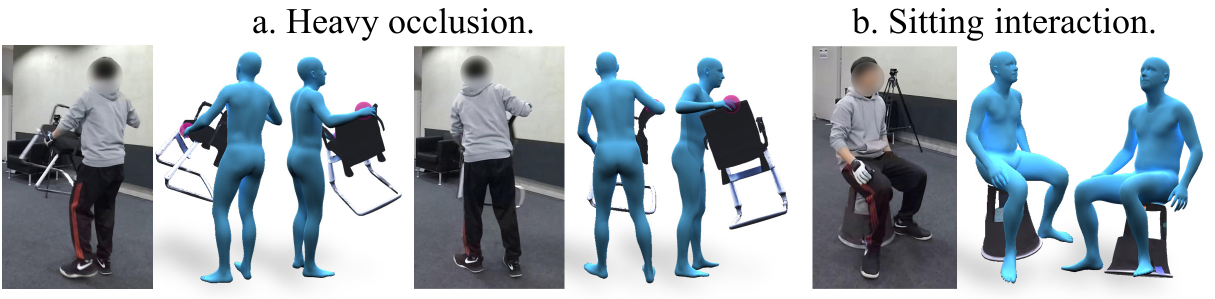}
    \caption{Additional qualitative examples.}
    \label{fig:rebuttal-example}
\end{figure}

\noindent\textbf{Interactions other than hand-objects}. We did evaluate the performance other than hand only interactions as BEHAVE test set also includes sitting, leaning, shoulder carrying. We classify sequences based on interaction types and evaluate the performance in the table below. We can see that errors are similar among different interaction types, demonstrating the robustness of our method. We also show one sitting example in \cref{fig:rebuttal-example}b.

\vspace{-0.2cm}
\begin{table}[ht]
  \centering
  \vspace{-5pt}
  \scriptsize
  \begin{tabular}{
    l|
    S[table-format=2.2,table-column-width=0.5cm]
    S[table-format=4.2,table-column-width=0.6cm]
    S[table-format=3.2,table-column-width=0.6cm]
    S[table-format=1.2,table-column-width=0.5cm]
    S[table-format=1.2,table-column-width=0.5cm]
    S[table-format=1.2,table-column-width=0.5cm]
  }
    Interaction type &
    {\tiny CD-h$\downarrow$} &
    {\tiny CD-o$\downarrow$} &
    {\tiny CD-c$\downarrow$} &
    {\tiny Acc-h$\downarrow$} &
    {\tiny Acc-o$\downarrow$} & 
    {\tiny Contact$\uparrow$} \\
    \hline
    Hand mainly  & 8.76  & 11.71   & 7.12   & 1.06 & 0.42 & 0.93\\
    Sitting       & 8.16 & 10.10 & 7.23 & 1.01 & 0.32 & 0.95\\
    Shoulder contact     & 10.24  & 12.39   & 9.27  & 1.03 & 0.48 & 0.95\\
    Leaning     & 9.37 &  11.44  & 8.28  & 1.05 & 0.31 & 0.93\\
    
  \end{tabular}
  \caption{Performance of our method per interaction type.}
  \label{tab:interaction-type}
\end{table}

\section{Limitation and Future Work }\label{sec:limitation}
\begin{figure}[ht]
    \centering
    \includegraphics[width=1.0\linewidth]{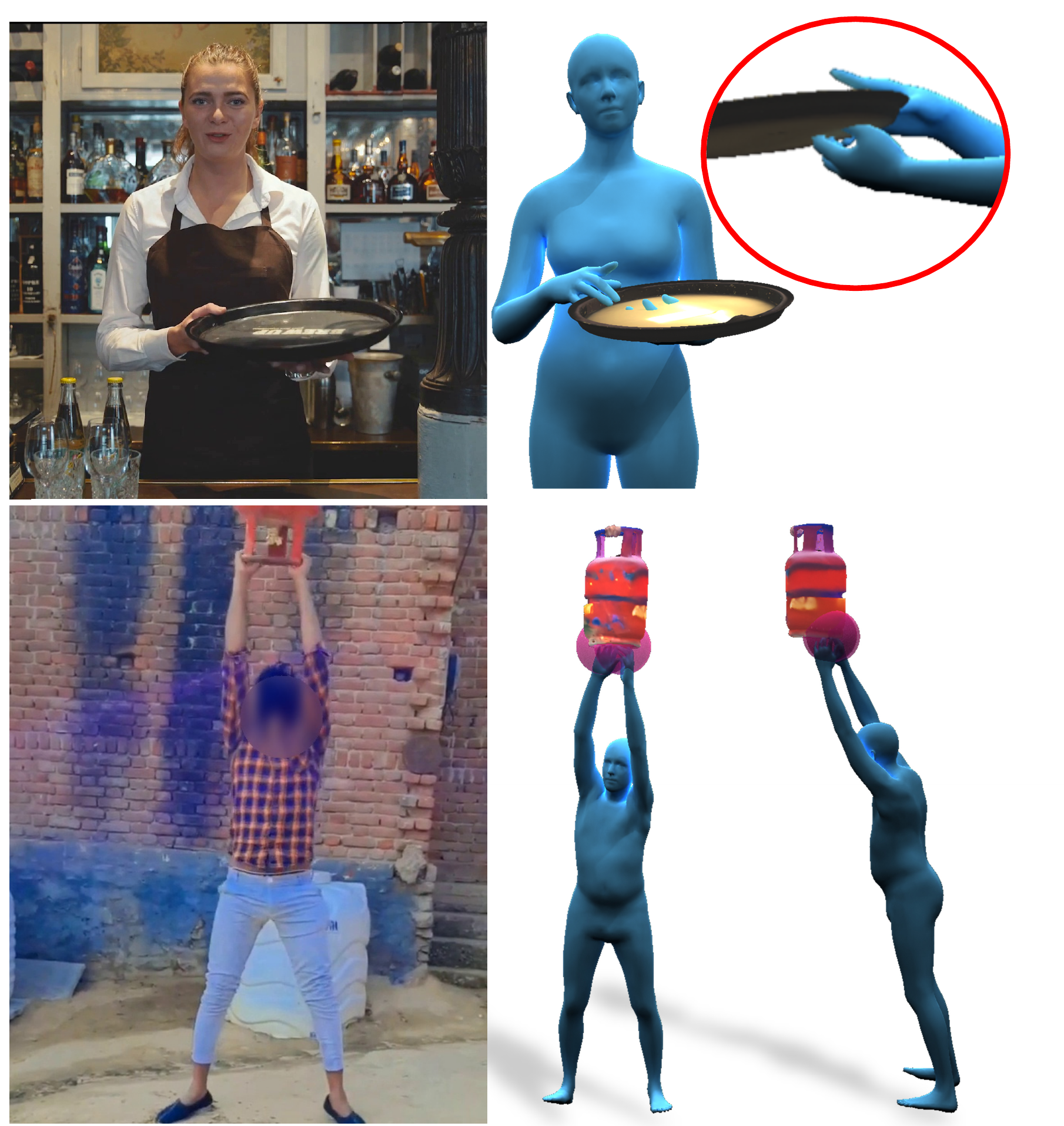}
    \caption{\textbf{Failure case examples.} Our method focuses on full body interaction and the detailed hand poses are not handled, which can be important for fine-grained object manipulation task (top row). Our method thus failed to reconstruct realistic finger poses for holding the plate. Under highly dynamic motion and extreme occlusion (bottom row), FoundationPose predicts flipped object pose for initialization. Such large rotation error is not able to be corrected by our refinement process in subsequent steps, leading to inaccurate reconstruction in the end. }
    \label{fig:failure}
\end{figure}
As the first step towards category agnostic 4D interaction reconstruction, our method shows strong generalization performance to in-the-wild videos, yet there are still some limitations. We show two typical failure cases in \cref{fig:failure}.

First, our method primarily targets full-body human-object interaction; consequently, it does not explicitly regress detailed finger articulatons. This limitation becomes particularly pronounced during interactions involving small-scale objects or those requiring fine-grained manipulation, such as grasping plates (\cref{fig:failure}, top row). In such scenarios, the absence of precise finger kinematics results in physically implausible interaction configurations, even when the full body pose is accurate. To bridge this gap, future iterations could integrate specialized hand pose estimators~\cite{pavlakos2024hand, zhang2025hawor, potamias2024wilor}. By solving for the hand parameters separately and fusing them with the full-body kinematic chain via optimization, one could achieve a holistic reconstruction that respects both macro-level body dynamics and micro-level contact physics.

Second, our method relies on FoundationPose~\cite{foundationposewen2024} for object pose initialization, subsequently refining these estimates using human interaction cues and visual evidence. A notable dependency bottleneck arises when the initializer fails significantly. FoundationPose can occasionally predict flipped 180-degree poses under conditions of rapid motion or severe occlusion (\cref{fig:failure}, bottom row). In these instances, the error magnitude is often too large for our refinement network to correct. A promising direction is the incorporation of temporal priors and motion infilling, similar to strategies employed in VisTracker~\cite{xie2023vistracker} or GLAMR~\cite{yuan2022glamr}. By leveraging information from visible frames to hallucinate motion in occluded segments, we can enforce temporal smoothness and recover from initialization failures.

\end{document}